# Repainting and Imitating Learning for Lane Detection

Yue He*
heyue04@baidu.com
Baidu Inc.
Beijing, China

Minyue Jiang*
jiangminyue@baidu.com
Baidu Inc.
Beijing, China

Xiaoqing Ye*
yexiaoqing@baidu.com
Baidu Inc.
Shanghai, China

Liang Du*
duliang@mail.ustc.edu.cn
Fudan University
Shanghai, China

Zhikang Zou
zouzhikang@baidu.com
Baidu Inc.
Shenzhen, China

Wei Zhang
zhangwei99@baidu.com
Baidu Inc.
Shenzhen, China

Xiao Tan
tanxiao01@baidu.com
Baidu Inc.
Shenzhen, China

Errui Ding
dingerrui@baidu.com
Baidu Inc.
Beijing, China

## ABSTRACT

Current lane detection methods are struggling with the invisibility lane issue caused by heavy shadows, severe road mark degradation, and serious vehicle occlusion. As a result, discriminative lane features can be barely learned by the network despite elaborate designs due to the inherent invisibility of lanes in the wild. In this paper, we target at finding an enhanced feature space where the lane features are distinctive while maintaining a similar distribution of lanes in the wild. To achieve this, we propose a novel Repainting and Imitating Learning (RIL) framework containing a pair of teacher and student without any extra data or extra laborious labeling. Specifically, in the repainting step, an enhanced ideal virtual lane dataset is built in which only the lane regions are repainted while non-lane regions are kept unchanged, maintaining the similar distribution of lanes in the wild. The teacher model learns enhanced discriminative representation based on the virtual data and serves as the guidance for a student model to imitate. In the imitating learning step, through the scale-fusing distillation module, the student network is encouraged to generate features that mimic the teacher model both on the same scale and cross scales. Furthermore, the coupled adversarial module builds the bridge to connect not only teacher and student models but also virtual and real data, adjusting the imitating learning process dynamically. Note that our method introduces no extra time cost during inference and can be plug-and-play in various cutting-edge lane detection networks. Experimental results prove the effectiveness of the RIL framework both on CULane and TuSimple for four modern lane detection methods. The code and model will be available soon.

## 1 INTRODUCTION

Lane detection is a crucial task in autonomous driving [5], which could serve as visual cues for advanced driver assistance systems (ADAS) to keep vehicles stably following lane markings. Thus, autonomous vehicles need to locate each lane's precise position.

With the development of deep learning, neural networks [32] have been widely used in lane detection for their compelling performance. Early deep-learning-based methods detect lanes through pixel-wise segmentation based framework [22, 23, 27], where each pixel is assigned a binary label to indicate whether it belongs to a lane or not. More recently, various anchor-based methods [3, 13, 30] are proposed, different forms of anchors such as line anchors and box anchors are used among these methods, aiming to let the networks focus the optimization on the line shape by regressing the relative coordinates. Besides, row-wise classification methods [14, 24, 26, 37] rely on the shape prior of lanes and predict the location for each row. Parametric prediction methods [15, 31] directly output parameters of curve equation for lines. Multi-task learning is further combined to improve lane detection accuracy in a complicated environment. For example, VPGNet [12] integrates road marking detection and vanishing point prediction to obtain auxiliary information for lane detection. However, additional manual labeling is time-consuming and laborious.

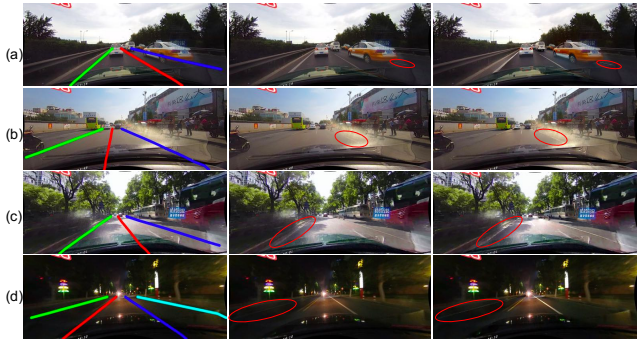

**Figure 1: Samples of complex scenarios in the CULane dataset: (a) Crowded (b) Dazzle (c) Shadow (d) Night. The first column shows the ground truth, and we highlight the elliptical region for comparison between lines before repainting (second column) and after repainting (third column).**

Either combining shape prior of lanes or designing the auxiliary task has shown its comprehensive consideration and competitive results for lane detection, it is still a challenging task due to many factors such as the wide variety of lane markings appearance including solid or dashed, white or yellow. Moreover, complex road and light conditions leading to occlusion and low illumination increase the invisibility of lanes. All these difficulties require the method to have the ability to extract lanes under complicated environments. Taking CULane dataset [23] as an example in Fig. 1, we present four representative scenarios, including crowded, dazzle, shadow, and night. In these scenarios, complete lanes cannot be visually detected by comparing lanes in the wild (the second column) with the ground truth annotations (the first column). The inherent invisibility of lanes hinders the progress of the algorithms. In this paper, we focus on finding a more discriminative lane feature space and maintaining a similar distribution of lanes in the wild. The Repainting and Imitating Learning (RIL) framework is proposed to increase the visibility of lanes by repainting module without extra data or labor labeling, and improve feature discrimination while transferring lane knowledge from teacher to student via imitating learning simultaneously.

First and foremost, the inherent invisibility of lanes in the wild intractable by current lane detection methods is alleviated by a simple but effective repainting module in our repainting step. Through this module, virtual data is generated based on the location of lanes annotated in the ground truth without the requirement of extra data or laborious labeling. This module highlights the fuzzy lanes and makes lines more prominent and continuous. As shown in Fig. 1, the lane regions in the third column become more distinctive and continuous while maintaining other non-lane regions unchanged. Based on these ideal lanes, the teacher will be trained in advanced and achieves upper bound performance.

A simple solution by directly adding the virtual data as data augmentation does not work, since there exists a distribution gap between virtual data and real data. Thus, to better utilize virtual data, imitating learning step is introduced including a scale-fusing distillation module and a coupled adversarial module. In the scale-fusing distillation module, different stages of feature representations from the teacher are treated as the ideal enhanced feature space. Noticing that the teacher model has the same architecture as the student. The teacher's feature maps of the same size as the student's feature maps are distilled directly. Besides, the teacher's larger feature maps are down-sampled to distill student's semantic feature maps simultaneously for finer lane details. Both the same scale and cross scale information are distilled, helping student to imitate the teacher. To further eliminate distribution gaps not only between different networks but also between different input data, coupled adversarial module is proposed to build a bridge to connect networks as well as data. A pair of discriminators are coupled by adding in another student's output of virtual data. The first net-sensitive discriminator is to distinguish teacher and student networks when they are fed into virtual data. The second data-sensitive discriminator is to distinguish between virtual data and real data which feed to the student network. By coupling two discriminators, the student can better imitate the enhanced teacher features dynamically through the learning process.

Our main contributions are summarized as follows:

- We introduce a simple yet effective Repainting and Imitating framework (RIL) for lane detection, focusing on discriminating lane features and maintaining the similar distribution of lanes in the wild by finding an enhanced feature space.
- We repaint the real lane data to ideal virtual data in the repainting step, achieving enhanced representation under complicated environments.
- We combine the scale-fusing distillation module with the coupled adversarial module in the imitating step, building the bridge between networks and data to weaken the learning gap.

The proposed RIL framework can be easily plug-and-play in most cutting-edge methods without any extra inference cost. Experimental results prove the effectiveness of RIL framework both on CULane [23] and TuSimple [33] for four modern lane detection methods including UFAST [26], ERFNet [27], ESA [11] and CondLaneNet [14] respectively.

## 2 RELATED WORK

**Lane detection** Recent approaches [4, 16, 19, 25] focus on the deep neural networks and significantly boost the lane detection performance due to the powerful representation learning ability. Some methods[11, 23, 27] treat lane detection as a semantic segmentation task. For instance, SCNN [23] designs slice-by-slice convolutions within feature maps to exchange pixel information between pixels across rows and columns in a layer. Inspired by network architecture search (NAS), CurveLanes-NAS [36] designs a lane-sensitive architecture to incorporate both long-ranged coherent lane information and short-ranged local lane information. Despite the promising results, the computational complexity in these methods brings heavy inference overhead. Therefore, row-wise classification based methods [9, 14, 26, 37] have been proposed for efficient lane detection. These approaches divide the input image into grids and predict the most probable cell to contain a part of a lane, which realize the trade-off between speed and accuracy. More recently, SGNet [28] introduces a structure-guided framework to accurately classify, locate and restore the shape of unlimited lanes. Unlike the systematic approach, we proposed RIL framework can be plug-and-play in various cutting-edge lane detection networks.

**Knowledge distillation** Knowledge distillation (KD) aims to transfer knowledge from a deep and complex teacher model to a shallower and simpler student model to facilitate supervised learning. Knowledge distillation has been applied in many vision tasks, such as object detection [2, 34], semantic segmentation [8, 17] and other visual tasks [35, 38]. Guided by smoother supervision of soft labels in the teacher model, the student could learn more hints on visual concepts and obtain a competitive or even a superior performance accompanied by a decrease in the computation complexity. Some works have studied knowledge distillation in lane detection [9, 10]. SAD [10] proposes a novel attention distillation approach to perform top-down and hierarchical knowledge distillation and gains prominent improvement without any additional supervision. IntRA-KD [9] exploits the application of inter-region affinity knowledge distillation for its simplicity and generality. Through building inter-region affinity graphs, the student learns to establish pairwise relationships between nodes and extracts structural knowledge from the teacher network. Different from previous works that use heavy teacher networks for knowledge distillation, we propose a simple yet effective sibling network with the same architecture that fundamentally exploits the feature association ability via the virtual-to-real feature distillation.

**Adversarial machine learning** [21] mention models have become unreliable as they may change the predictions even yield incorrect predictions when encountered with input data added with an 'adversarial noise' [29]. In recent years, adversarial attacks have been explored in computer vision research, aiming to find a minimal perturbation that maximizes the risk of the model making wrong predictions. Existing defense mechanisms [1, 7, 18] against adversarial attacks [6, 20] attempt to reduce the impact of adversarial examples. Building adversarial defense mechanisms can be done in two ways, either by directly modifying the classifier to make the system more robust or by transforming adversarial examples in inference time. Adversarial learning methods can also be treated as a special regularization for machine learning which discriminates data whether in-distribution or out-distribution. In this work, we use adversarial learning to build the bridge between the different distribution of networks' parameters and data. The classifier of defect inspection in [39] distinguishes synthesized samples from real ones explicitly, which is similar to our single adversarial module design. In addition, our coupled adversarial module contains another discriminator which can make the learning process more effective and build the bridge not only between different domain data but also between different networks.

## 3 METHODOLOGY

In this section, we first briefly introduce the overview framework of our proposed RIL and then illustrate each module in RIL. Note that our method can be flexibly combined with many cutting-edge lane detection methods such as [11, 14, 26, 27]. In this section, we adapt ERFNet [27] to RIL framework for a demonstration.

### 3.1 Framework overview

As shown in Fig. 2, our proposed RIL framework essentially follows the teacher-student knowledge distillation paradigm while containing three unique modules: (a) Repainting module that generates the virtual training data via repainting the lane regions while keeping non-lane regions unchanged; (b) Scale-fusing distillation module distills enhanced lane features from the teacher to the student via same scale and cross scales; (c) Coupled adversarial module which dynamically discriminates three outputs: teacher's output of virtual scene, student's output of virtual scene and real scene.

The training process of RIL contains two major steps: the repainting step and the imitating step. In the repainting step, we firstly use the repainting module to map real dataset $R = \{r_i, y_i\}_{i=1}^{N}$ to virtual dataset $V = \{v_i, y_i\}_{i=1}^{N}$, where $N$ is the number of training images, $y_i$ is the binary lane mask of ground truth label for $i$th real image $r_i$ and virtual image $v_i$. Through such simple repainting operations, lanes with low visibility are effectively enhanced. Secondly, the teacher network is trained with the virtual data which can achieve enhanced performance and reaches the upper bound of datasets such as CULane. In the following imitating learning step, the scale-fusing distillation module and coupled adversarial module are added to assist in transferring knowledge from the frozen teacher network to the training student network. During the inference phase, only the student remains to compute the feature map for lane detection, taking no extra computation cost.

In the subsequent sections, we provide detailed illustrations of each module in our proposed RIL.

### 3.2 Repainting module

As shown in Fig. 3, the repainting module is designed to enhance the original image for more clear and crisp lanes. Concretely, the virtual data generation process in our proposed repainting module contains three steps for each sample: (1) The lane regions are extracted according to the ground truth (GT) mask $y_i$ of the $i$th image $r_i$ from real dataset $R$. (2) For each lane region, we apply image contrast enhancement method to enhance the visibility. Here we practically choose the simple linear enhancement since we have experimented with different contrast enhancement methods while ended up with almost the same lane detection results. (3) Finally, as a common post processing operation for regional image editing, we also adopt poisson fusion to achieve a more realistic composition for the enhanced lane region and the background pixels. Formally, we set lane region as selection $\Omega$ and source image as $g$, the $f^*$ is our destination. The poisson equation as $\triangle f = \triangle g$ over $\Omega$, with $f|_{\partial\Omega} = f^*|_{\partial\Omega}$ is solved in each color channel, where $\triangle * = \frac{\partial^2 *}{\partial x^2} + \frac{\partial^2 *}{\partial y^2}$ is the Laplacian operator and $f$ is an unknown scalar function defined over the interior of $\Omega$.

As a result, the generated virtual data can be seen in Fig. 1 and Fig. 3, the repainting module seamlessly strengthens the lane regions while maintaining other non-lane regions unchanged. Taking the advantage of the virtual dataset $V$, the teacher network can be trained in advance to meet the requirement of the subsequent distillation module and adversarial module. The advantage of this lane region enhancement for the real data is two folds: On the one hand, the increasing contrast between lane regions and non-lane regions makes the teacher network easily capture locations of lanes to achieve enhanced performance, building a more discriminative lane feature space. On the other hand, the modification reduces the feature gap between teacher and student in the following scale-fusing distillation module, helping the student pay more attention to lane regions while maintaining the similar distribution of lanes in the wild.

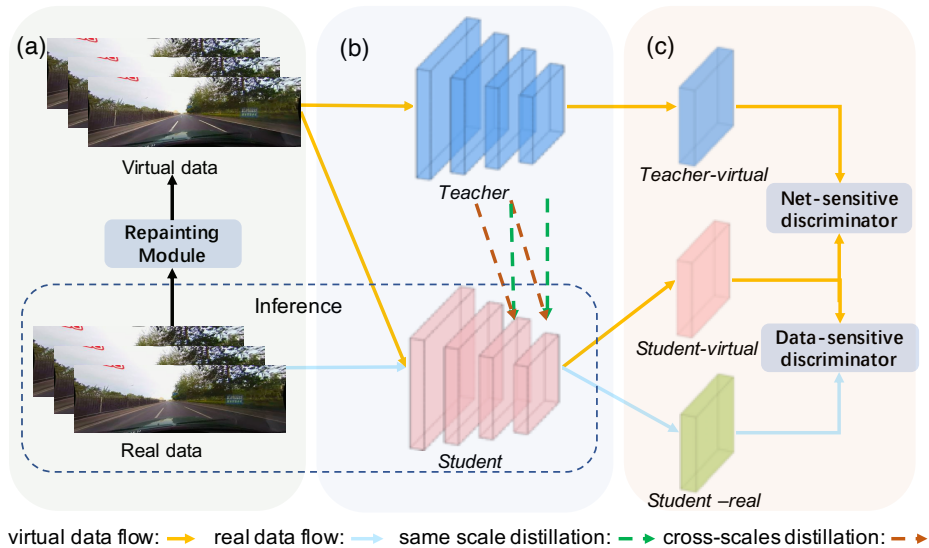

Figure 2: The overview of RIL for lane detection. (a) The repainting module is responsible for building a virtual dataset based on real images to highlight lanes region. (b) The scale-fusing distillation module distills the teacher's virtual features to the student. The green and red dashed lines represent distillation at the same scale and at cross scales respectively. (c) Subsequently, a coupled adversarial module with $D_{net}$ and $D_{data}$ is proposed, which not only distinguishes teacher-virtual and student-virtual features come from teacher model and student model, but also distinguishes student-virtual and student-real features from virtual data and real data, thus establishing a dynamic bridge.

### 3.3 Scale-fusing distillation module

Traditional knowledge distillation frameworks encounter a great challenge for lane segmentation models due to the inherent slender characteristics of lanes usually leads to fuzzier features as the model layer stacks. To address this, we propose scale-fusing distillation mechanism in which layers in the student learn to transfer knowledge not only from the same layer but also from the shallower layer with finer-detailed features of the teacher. Besides, since the teacher model takes enhanced images as input which produces clear and strong features for lane regions while the student model intends to recover enhanced features with only the original image as input during the inference. Such a domain gap between the inputs of teacher and student increases the difficulty of distillation. We further propose to perform cross domain distillation between the teacher and student and the self-distillation for the student simultaneously.

Specifically, the goal of our distillation is to make student backbone's output approximately teacher backbone's output which is $G(r_i, \theta_{stu}, j) \simeq G(v_i, \theta_{tea}, j)$. Due to the domain gap between virtual data and lanes in the wild, we further introduce virtual data fed to the student network and our goal of the distillation will be disassembled into two parts, one is $G(v_i, \theta_{tea}, j) \simeq G(v_i, \theta_{stu}, j)$, the other is $G(v_i, \theta_{tea}, j) \simeq G(r_i, \theta_{stu}, j)$. That is to say, virtual data is fed to both the teacher network and student network, and real data is only fed to the student network. Feature maps of different scales are generated which can be written as follows:

$$T_{v_i}^j = G(v_i, \theta_{tea}, j), \quad (1)$$

$$S_{r_i}^j = G(r_i, \theta_{stu}, j), \quad (2)$$

$$S_{v_i}^j = G(v_i, \theta_{stu}, j), \quad (3)$$

where $T_{v_i}^j$, $S_{v_i}^j$ and $S_{r_i}^j$ indicate the feature map of $j$th stage generated by the corresponding data and network, $\theta_{tea}$, $\theta_{stu}$ means the network parameters of teacher and student respectively, $G(*)$ represents the backbone part of the inference process.

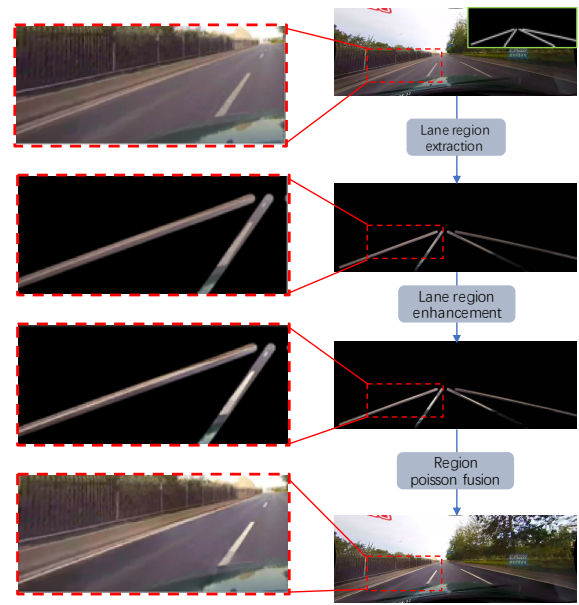

Figure 3: Illustration of the repainting module. The left column is a partial enlargement of the image in the flowchart on the right, and the green box shows ground truth of the lane region.

Formally, we exploit $L2$ distance as the loss function to minimize the multiple same scale feature distances:

$$\mathcal{L}_{distill}^{same} = \frac{1}{N_{same}} \sum_{j=1}^{N_{same}} (||T_{v_i}^j - S_{v_i}^j||^2 + ||T_{v_i}^j - S_{r_i}^j||^2), \quad (4)$$

where $N_{same}$ indicates the number of same scale distillation, and we use $N_{same} = 2$ in all experiments. Meanwhile, we also add the cross-scale distillation to strengthen the finer details of lane regions from teacher finer feature maps:

$$\mathcal{L}_{distill}^{cross} = \frac{1}{N_{cross}} \sum_{j=1}^{N_{cross}} (||A(T_{v_i}^{j-1}) - S_{v_i}^j||^2 + ||A(T_{v_i}^{j-1}) - S_{r_i}^j||^2), \quad (5)$$

where $N_{cross}$ indicates the number of cross scale distillations and we also use $N_{cross} = 2$ in all experiments, $A(*)$ represents the down-sample alignment operation. Teacher's outputs are down-sampled to generate the same aligned scale feature map of student's outputs to compute the cross-scale distillation loss.

Finally, the total distillation loss function is the sum of the same scale distillation loss and cross scale distillation loss:

$$\mathcal{L}_{distill} = \mathcal{L}_{distill}^{same} + \mathcal{L}_{distill}^{cross}. \quad (6)$$

### 3.4 Coupled adversarial module

In fact, there are naturally two gaps: not only between the virtual data $v_i$ and the real data $r_i$ introduced by the repainting module, but also between the learned parameters of the frozen teacher network $\theta_{tea}$ and the student network $\theta_{stu}$. Thus, merely adopting the scale-fusing distillation cannot compensate the distribution gap well with the fixed $L2$ distance loss function. In order to further reduce the gaps, we design the coupled adversarial module by adding $S_{v_i}^j = G(r_i, \theta_{stu}, j)$ to build the bridge between $T_{v_i}^j$ and $S_{r_i}^j$, dynamically adjusting the transfer learning procedure.

Compared to the traditional single discriminator adversarial module $D_{single}$ which builds the adversarial module between $T_{v_i}^j$ and $S_{r_i}^j$, the proposed coupled adversarial module follows the rule of the triangle: the combination of two discriminator ($D_{net}$ and $D_{data}$) is greater than the other discriminator ($D_{single}$), by adding the bridge of $S_{v_i}^j$. Though the training process is slightly more complicated, it achieves better performance.

In detail, a pair of discriminators including net-sensitive discriminator $D_{net}$ and data-sensitive discriminator $D_{data}$ are coupled, discriminating three different type of feature maps $T_{v_i}^j$, $S_{v_i}^j$ and $S_{r_i}^j$. By using net-sensitive discriminator $D_{net}$ to discriminate two feature maps $T_{v_i}^j$ and $S_{v_i}^j$, data discrepancy is compensated, and the student network focus on minimizing the parameter discrepancy during the first part of adversarial learning. By using data-sensitive discriminator $D_{data}$ to discriminate the latter two feature maps $S_{v_i}$ and $S_{r_i}$, parameter discrepancy is compensated, and the student network focus on minimizing the data discrepancy during the second part of adversarial learning. The student targets to minimize cross entropy loss, formally:

$$\mathcal{L}_{adv}^{net} = -(\log(D_{net}(T_{v_i}^j)) + \log(1 - D_{net}(S_{v_i}^j))), \quad (7)$$

$$\mathcal{L}_{adv}^{data} = -(\log(D_{data}(S_{v_i}^j)) + \log(1 - D_{data}(S_{r_i}^j))), \quad (8)$$

and the final adversarial loss function for student is the sum of:

$$\mathcal{L}_{adv} = \mathcal{L}_{adv}^{net} + \mathcal{L}_{adv}^{data}. \quad (9)$$

### 3.5 Loss functions

Combining the original lane detection loss, scale-fusing distillation loss and adversarial loss together, the overall objective loss function to minimize is formulated as follows:

$$\mathcal{L}_{total} = \mathcal{L}_{lane} + \mathcal{L}_{distill} + \mathcal{L}_{adv}, \quad (10)$$

where $L_{lane}$ and $L_{distill}$ can ensure that student is closer to teacher and $L_{adv}$ is to balance the imitating process in the presence of gaps between different networks and different data. In conclusion, scale-fusing distillation module and coupled adversarial module are only conducted during training, thus no extra time cost during inference and can be plug-and-play in various modern lane detection methods.

## 4 EXPERIMENTS

### 4.1 Dataset

We conduct extensive experiments on two widely used lane detection benchmark datasets: the CULane dataset [23] and the TuSimple dataset [33]. An overview of the datasets can be seen in Table 1.

Table 1: Details of the CULane and TuSimple datasets.

| Dataset  | Train | Validation | Test  | Road type         |
| -------- | ----- | ---------- | ----- | ----------------- |
| CULane   | 88.9K | 9.7K       | 34.7K | Urban and Highway |
| TuSimple | 3.3K  | 0.4K       | 2.8K  | Highway           |

**The CULane dataset.** The CULane [23] dataset consists of more than 55 hours of videos which comprises urban and highway scenarios. It comes from nine different scenarios, including normal, crowd, curve, dazzle night, night, no line, and arrow in the urban areas and highways.

**The TuSimple dataset.** TuSimple[33] dataset is collected with stable lighting conditions in highways, which contains 3268/358/2782 for train/validation/test respectively.

### 4.2 Implementation details

**Network architecture.** Our RIL framework can be plug-and-play with many modern lane detection methods including segmentation based and row-wise classification based methods. In detail, we implement our method with a lightweight row-wise classification framework such as UFast [26] and segmentation based methods

ERFNet [27]/ESA [11], and current state-of-the-art row-wise classification framework CondLaneNet [14].

**Training details.** For UFast [26], the numbers of row anchors are defined according to the image height of the dataset. Specifically, the row anchors distribution of the CULane dataset range from 260 to 530 with a step of 10 for its 540 image height. The counterpart of the TuSimple dataset ranges from 160 to 710 as the image height is 720. The number of gridding cells is set to 150 and 100 for the CULane and the Tusimple dataset respectively.

For ERFNet [27], our lane detection model on CULane is trained with 12 epochs, with 12 images per batch and each image is resized to 976x208 after ignoring the top 240 pixels on the image. We use SGD to train our models and the initial learning rate is set to 0.01. Since there is no default configuration on TuSimple, we resize the images to 368×640, set the initial learning rate to 0.02, and total training epochs to 60.

For ESA [11], the images of CULane and TuSimple are resized to 288 × 800 and 368×640, respectively. Moreover, SGD is used as the optimizer, and the initial learning rate and batch size are set to 0.1 and 12, respectively.

For Condlanenet [14], input images are resized to 800x320 pixels during training and testing. In the optimizing process, we use Adam optimizer and step learning rate decay method with an initial learning rate of 3e-4. We train 16 and 70 epochs for CULane and TuSimple with a batch size of 32 respectively.

### 4.3 Evaluation metrics

**The CULane dataset.** Intersection-over-union (IoU) is calculated between predictions and ground truth where each lane is treated as a 30-pixel-width line. Predicted lanes whose IoU are larger than a threshold (0.5) are considered true positives (TP). In addition, the $F_1$-measure is used as an evaluation metric and is defined as follows:

$$F_1 = \frac{2 \times Precision \times Recall}{Precision + Recall}, \quad (11)$$

where $Precision = \frac{TP}{TP+FP}$, $Recall = \frac{TP}{TP+FN}$.

**The TuSimple dataset.** For TuSimple dataset, both $F_1$-measure and accuracy are used to evaluate. The definition of $F_1$ score is similar in Eq 11 and the format of accuracy is:

$$accuracy = \frac{\Sigma_{clip} C_{clip}}{\Sigma_{clip} S_{clip}}, \quad (12)$$

in which $C_{clip}$ is the number of lane points predicted correctly (mismatch distance between prediction and ground truth is within a certain range) and $S_{clip}$ is the total number of ground truth points in each *clip*. We also evaluate the rate of false positive (FP) and false negative (FN) on prediction results. For example, lane with accuracy greater than 85% is considered a true positive otherwise false positive or false negative.

### 4.4 Quantitative results

To verify the effectiveness of our proposed RIL framework, we combine it with multiple algorithms including UFast [26], ERFNet [27], ESA [11], and CondLaneNet [14]. As illustrated in Table 2 and Table 3, the proposed method achieves a competitive performance on the CULane of 79.72% $F_1$ score, and on the TuSimple of 97.26% $F_1$ score. In some experiments, we did not reproduce successfully the results in the original papers, thus we report our reproducing results marked with *.

As shown in Table 2, RIL can consistently improve the performance under various configurations on the CULane dataset. Especially, for the sub-category of shadow lanes, RIL can achieve 2.57% higher $F_1$ score than the original ERFNet. We can also achieve improvement on the TuSimple dataset similarly. However, compared with the CULane dataset, the improvement on the TuSimple is smaller. We analyze that it is due to the less obscured and blurred lanes in the highway scenario than in the urban scenario, the performance of original models already reaches a plateau, the repainting module cannot generate more informative lane regions and the teacher network has a similar performance to the student.

We visualize lane detection results on the CULane(top three rows), and TuSimple(bottom three rows), the results of UFast-R34, ESA, and CondLaneNet-L are shown in Figure 4. Meanwhile, with the assistance of RIL, the prediction of lane regions are more continuous than the corresponding original methods and solved some missed detection problems on the sides of the image where the lanes are not visible clearly.

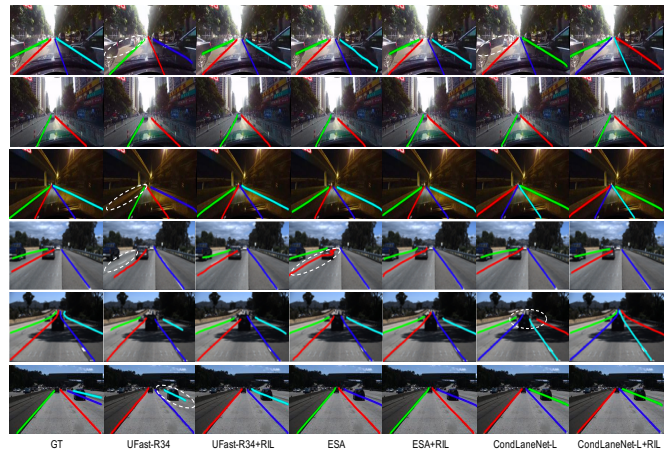

**Figure 4: Visualization results of six images from the CULane(top three rows) and TuSimple(bottom three rows) test dataset separately. Different columns represent the results of different methods: GT, UFast-R34 and UFast-R34+RIL, ESA and ESA+RIL, CondLaneNet-L and CondLaneNet-L+RIL, where the effect of the white circle position has been improved by our RIL.**

### 4.5 Ablation study

We performed ablation experiments on the Culane dataset based on the ERFNet [27]. The experiments are all conducted in the same settings as described in Sec. 4.2 if not specially mentioned.

**Performance of different networks on virtual and real data.**

Table 4 summarizes the performance of different networks based on virtual and real data. The baseline performance is 73.10% $F_1$ score. The teacher network can achieve an oracle performance of 99.85% $F_1$ score with the virtual data, laying the foundation for the following distillation module and adversarial module. Combining all three modules, the student network improves 1.23% $F_1$ score comparing the baseline on the real data.

Table 2: Comparison of F1 measure of state-of-the-art methods on the CULane dataset. For cross road, only FP is shown. +RIL means adding RIL based on the existing model where shading is gray and results reproduced by our experiments are marked with *.

| Method Based | Method | Total | Normal | Crowded | Dazzle | Shadow | No line | Arrow | Curve | Cross | Night | FPS | GFlops |
|---|---|---|---|---|---|---|---|---|---|---|---|---|---|
| UFAST (lightweight) | UFAST-Res18 [26] | 68.40 | 87.70 | 66.00 | 58.40 | 62.80 | 40.20 | 81.00 | 57.90 | 1743 | 62.10 | 322.5 | - |
| | **UFAST-Res18+RIL** | **71.35** | **90.21** | **68.93** | **59.25** | **63.68** | **41.89** | **84.62** | **60.11** | **1577** | **66.16** | **322.5** | - |
| | UFAST-Res34 [26]* | 70.36 | 89.10 | 68.47 | 56.98 | 68.06 | 42.93 | 84.63 | 61.84 | 2037 | 64.47 | 175.0 | - |
| | **UFAST-Res34+RIL** | **72.22** | **90.32** | **69.92** | **59.31** | **70.40** | **42.53** | **85.80** | **64.34** | **1729** | **66.83** | **175.0** | - |
| ERFNet (widely-used) | ERFNet [27] | 73.10 | 91.50 | 71.60 | 66.00 | 71.30 | 45.10 | 87.20 | 66.30 | 2199 | 67.10 | - | - |
| | **ERFNet+RIL** | **74.33** | **91.78** | **72.83** | **64.41** | **73.87** | **47.57** | **87.28** | **67.69** | **2222** | **68.75** | - | - |
| | ESA [11] | 74.20 | 92.00 | 73.10 | 63.10 | 75.10 | 45.80 | 88.10 | 68.80 | 2001 | 69.50 | - | - |
| | **ESA+RIL** | **75.01** | **92.09** | **73.16** | **63.98** | **77.16** | **48.23** | **87.84** | **67.83** | **2007** | **70.30** | - | - |
| CondLaneNet (SOTA) | CondLaneNet-S [14] | 78.14 | 92.87 | 75.79 | 70.72 | 80.01 | 52.39 | 89.37 | 72.40 | 1364 | 73.23 | 220 | 10.2 |
| | **CondLaneNet-S+RIL** | **78.67** | **93.18** | **76.93** | **70.27** | **80.37** | **52.22** | **89.19** | **74.80** | **1149** | **73.48** | **220** | **10.2** |
| | CondLaneNet-M [14] | 78.74 | 93.38 | 77.14 | 71.17 | 79.93 | 51.85 | 89.89 | 73.88 | 1387 | 73.92 | 152 | 19.6 |
| | **CondLaneNet-M+RIL** | **78.98** | **93.39** | **77.75** | **72.07** | **81.00** | **52.62** | **89.42** | **72.58** | **1342** | **73.94** | **152** | **19.6** |
| | CondLaneNet-L [14] | 79.48 | 93.47 | 77.44 | 70.93 | 80.91 | 54.13 | 90.16 | 75.21 | 1201 | 74.80 | 58 | 44.8 |
| | **CondLaneNet-L+RIL** | **79.72** | **93.60** | **78.58** | **70.30** | **81.09** | **54.38** | **89.76** | **74.82** | **1364** | **74.99** | **58** | **44.8** |

Table 3: Comparison of state-of-the-art methods on the TuSimple dataset. +RIL means adding RIL based on the existing model where shading is gray and results reproduced by our experiments are marked with *.

| Method Based | Method | F1 | Acc | FP | FN | FPS | GFlops |
|---|---|---|---|---|---|---|---|
| UFAST (lightweight) | UFAST-ResNet18 [26]* | 87.37 | **95.33** | 19.49 | 4.47 | 312.5 | - |
| | **UFAST-Res18+RIL** | **87.42** | 95.31 | 19.44 | 4.42 | 312.5 | - |
| | UFAST-Res34 [26]* | 87.23 | **95.39** | 19.63 | 4.62 | 169.5 | - |
| | **UFAST-Res34+RIL** | **87.47** | 95.38 | 19.39 | 4.37 | 169.5 | - |
| ERFNet (widely-used) | ERFNet [27]* | 95.18 | 94.90 | 3.79 | 5.63 | - | - |
| | **ERFNet+RIL** | **95.43** | **95.22** | 4.63 | 4.49 | - | - |
| | ESA [11]* | 95.93 | 94.90 | 3.75 | 4.37 | - | - |
| | **ESA+RIL** | **95.94** | **95.27** | 3.80 | 4.31 | - | - |
| CondLaneNet (SOTA) | CondLaneNet-S [14] | **97.01** | **95.48** | 2.18 | 3.80 | 220 | 10.2 |
| | **CondLaneNet-S+RIL** | 96.97 | 95.36 | 2.22 | 3.81 | 220 | 10.2 |
| | CondLaneNet-M [14] | 96.98 | 95.37 | 2.20 | 3.82 | 154 | 19.6 |
| | **CondLaneNet-M+RIL** | **97.15** | **95.55** | 2.08 | 3.60 | 154 | 19.6 |
| | CondLaneNet-L [14] | 97.24 | **96.54** | 2.01 | 3.50 | 58 | 44.8 |
| | **CondLaneNet-L+RIL** | **97.26** | 96.30 | 1.82 | 3.63 | 58 | 44.8 |

Table 4: The $F_1$ score performance of Teacher/Student model on Virtual/Real data.

| Networks | Virtual | Real |
|---|---|---|
| Baseline | 75.80 | 73.10 |
| Teacher | 99.85 | 0.53 |
| Student | 76.62 | 74.33 |

It is worth mentioning that the oracle performance of the teacher network indicates the distinctive feature extracted by this network. The improvement of performance by transferring knowledge from teacher to student is promising. However, there still exists a large gap between virtual data and real data base on the experiment that the teacher network has a poor performance on real data, thus we cannot directly use teacher network when inference. As the virtual data and real data feed to the student network simultaneously, it illustrates the effectiveness of the repainting module. What's more, with the bridge of $S_{v_i}^j$, transferring knowledge can be smoother due to the performance gap of $T_{v_i}^j$ and $S_{v_i}^j$ is smaller than gap of $T_{v_i}^j$ and $S_{r_i}^j$, which is also proven in the Table 7.

Fig. 5 visualizes the networks' features and the lane predictions applied to on CULane test data. The input images and the features and predictions of ERFNet baseline, teacher, and student are shown respectively. It is clear that the teacher's feature has a more distinctive feature representation. As a result, student of RIL has better lane particulars than the ERFNet baseline.

**Effectiveness of different modules in RIL.** This experiment evaluates the impact of the scale-fusing distillation module and the coupled adversarial module in Table 5, in which the ablation components are added one by one to measure their effectiveness. Adding the scale-fusing distillation module brings a performance boost of 0.84% $F_1$ score. Then combining coupled adversarial module with the scale-fusing distillation module brings a total performance improvement of 1.23% $F_1$ score, showing the complementarity of these two modules.

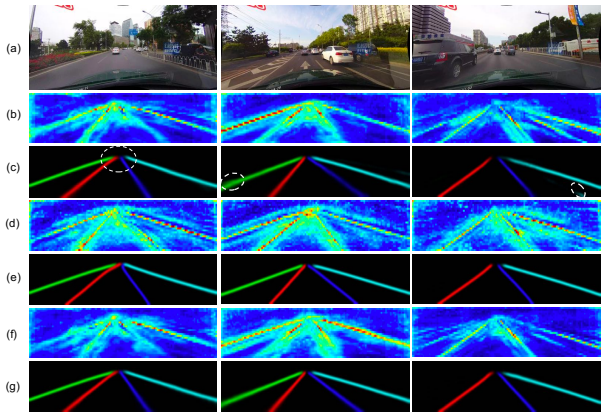

Figure 5: RIL backbone's features and the predictions on CULane test data, where the effect of the white circle position has been improved by our RIL. (a) is the input images. (b)(d)(f) are feature maps of baseline, teacher, and student. (c)(e)(g) are predictions of baseline, teacher, and student.

**Table 5: Ablation study of proposed modules base on the CULane dataset.**

| Scale-fusing distillation | Coupled adversarial | F1 |
|---|---|---|
|  |  | 73.10 |
| ✓ |  | 73.94(+0.84) |
| ✓ | ✓ | 74.33(+1.23) |

**Table 6: Comparison of different distillation scales for feature maps base on the CULane dataset.**

| Distillation Scale | F1 |
|---|---|
| w/o | 73.10 |
| Same scale | 73.67 |
| Scale-fusing | 73.94 |

**Different designs of distillation.** The designs of distillation between different feature map scales (the size of the feature map) are shown in Table 6. Compared to the baseline without distillation, we can find that the same scale feature distillation improves the $F_1$ score by 0.57%. Based on the same scale distillation method, we increase cross scale distillation, that is, the shallower scale finer features of the teacher are extracted to distill high-level small scale features of the student. This cross scale distillation brings another improvement of 0.27% of $F_1$ score. As a result, the combination of the same scale and cross scale is our default distillation choice.

**Single adversarial module vs. Coupled adversarial module.**
The most straightforward way to combine adversarial modules is by adding a single discriminator $D_{single}$ which discriminates $T_{v_i}^j$ and $S_{r_i}^j$ in which no virtual data is fed into student network. However, directly adding $D_{single}$ cannot bring an improvement based on scale-fusing distillation module. The reason behind that is the large gap between $T_{v_i}^j$ and $S_{r_i}^j$, discriminator can easily discriminate these two feature maps which fail to play its role during the training process. However, the coupled adversarial module meets the demand of the learning process by building the bridge of $S_{v_i}^j$. The net-sensitive discriminator $D_{net}$ only needs to discriminate parameters of the networks ($T_{v_i}^j$ and $S_{v_i}^j$) and the data-sensitive discriminator $D_{data}$ only need to discriminate different input data ($S_{v_i}^j$ and $S_{r_i}^j$). The narrow of the domain gaps can adjust the learning process dynamically according to the quality of feature maps from teacher and student. As shown in Table 7, coupled adversarial module can make another 0.39% improvement of $F_1$ score based on the scale-fusing distillation module.

**Table 7: Comparison of different adversarial learning on the CULane dataset. We add a single adversarial module and remove virtual data fed to the student for comparison which is more similar to the traditional distillation network.**

| Method | F1 |
|---|---|
| Scale-fusing distillation | 73.94 |
| Single adversarial | 73.90 |
| Coupled adversarial | 74.33 |

In short, the above experimental results prove that the repainting of the images effectively improves the distinctiveness of the teacher model's feature space while maintaining the similar distribution of lanes in the wild, and this enhanced teacher plays a guiding role for the student to imitate. Besides, the scale-fusing distillation module and the coupled adversarial module are complementary to each other. Thus we combine these two modules in imitating learning step.

## 5  CONCLUSION

In this paper, we propose a lane detection framework named Repainting and Imitating Learning (RIL), which contains a pair of teacher and student models without any extra data or extra laborious labeling with three components: the repainting module in the repainting step, scale-fusing distillation module and the coupled adversarial module in the imitating step. The proposed RIL framework can be directly plug-and-play in different methods without any extra cost at inference. Our method is evaluated on CULane and TuSimple and experimental results prove the effectiveness of our RIL framework. Since the current model has not achieved the effect of the teacher model, we will carry out follow-up work from the perspective of exploring the essence of the disparity between students and teachers.

# REFERENCES

[1] Anurag Arnab, Ondrej Miksik, and Philip HS Torr. 2018. On the robustness of semantic segmentation models to adversarial attacks. In *Proceedings of the IEEE Conference on Computer Vision and Pattern Recognition*. 888–897.

[2] Guobin Chen, Wongun Choi, Xiang Yu, Tony Han, and Manmohan Chandraker. 2017. Learning efficient object detection models with knowledge distillation. *Advances in neural information processing systems* 30 (2017).

[3] Zhenpeng Chen, Qianfei Liu, and Chenfan Lian. 2019. Pointlanenet: Efficient end-to-end cnns for accurate real-time lane detection. In *2019 IEEE intelligent vehicles symposium (IV)*. IEEE, 2563–2568.

[4] Zhe Ming Chng, Joseph Mun Hung Lew, and Jimmy Addison Lee. 2021. Roneld: Robust neural network output enhancement for active lane detection. In *2020 25th International Conference on Pattern Recognition (ICPR)*. IEEE, 6842–6849.

[5] Andreas Geiger, Philip Lenz, and Raquel Urtasun. 2012. Are we ready for autonomous driving? the kitti vision benchmark suite. In *2012 IEEE conference on computer vision and pattern recognition*. IEEE, 3354–3361.

[6] Ian J Goodfellow, Jonathon Shlens, and Christian Szegedy. 2014. Explaining and harnessing adversarial examples. *arXiv preprint arXiv:1412.6572* (2014).

[7] Chuan Guo, Mayank Rana, Moustapha Cisse, and Laurens Van Der Maaten. 2017. Countering adversarial images using input transformations. *arXiv preprint arXiv:1711.00117* (2017).

[8] Tong He, Chunhua Shen, Zhi Tian, Dong Gong, Changming Sun, and Youliang Yan. 2019. Knowledge adaptation for efficient semantic segmentation. In *Proceedings of the IEEE/CVF Conference on Computer Vision and Pattern Recognition*. 578–587.

[9] Yuenan Hou, Zheng Ma, Chunxiao Liu, Tak-Wai Hui, and Chen Change Loy. 2020. Inter-region affinity distillation for road marking segmentation. In *Proceedings of the IEEE/CVF Conference on Computer Vision and Pattern Recognition*. 12486–12495.

[10] Yuenan Hou, Zheng Ma, Chunxiao Liu, and Chen Change Loy. 2019. Learning lightweight lane detection cnns by self attention distillation. In *Proceedings of the IEEE/CVF international conference on computer vision*. 1013–1021.

[11] Minhyeok Lee, Junhyeop Lee, Dogyoon Lee, Woojin Kim, Sangwon Hwang, and Sangyoun Lee. 2022. Robust lane detection via expanded self attention. In *Proceedings of the IEEE/CVF Winter Conference on Applications of Computer Vision*. 533–542.

[12] Seokju Lee, Junsik Kim, Jae Shin Yoon, Seunghak Shin, Oleksandr Bailo, Namil Kim, Tae-Hee Lee, Hyun Seok Hong, Seung-Hoon Han, and In So Kweon. 2017. Vpgnet: Vanishing point guided network for lane and road marking detection and recognition. In *Proceedings of the IEEE international conference on computer vision*. 1947–1955.

[13] Xiang Li, Jun Li, Xiaolin Hu, and Jian Yang. 2019. Line-CNN: End-to-End Traffic line detection with line proposal unit. *IEEE Transactions on Intelligent Transportation Systems* 21, 1 (2019), 248–258.

[14] Lizhe Liu, Xiaohao Chen, Siyu Zhu, and Ping Tan. 2021. Condlanenet: a top-to-down lane detection framework based on conditional convolution. In *Proceedings of the IEEE/CVF International Conference on Computer Vision*. 3773–3782.

[15] Ruijin Liu, Zejian Yuan, Tie Liu, and Zhiliang Xiong. 2021. End-to-end lane shape prediction with transformers. In *Proceedings of the IEEE/CVF winter conference on applications of computer vision*. 3694–3702.

[16] Tong Liu, Zhaowei Chen, Yi Yang, Zehao Wu, and Haowei Li. 2020. Lane detection in low-light conditions using an efficient data enhancement: Light conditions style transfer. In *2020 IEEE Intelligent Vehicles Symposium (IV)*. IEEE, 1394–1399.

[17] Yifan Liu, Ke Chen, Chris Liu, Zengchang Qin, Zhenbo Luo, and Jingdong Wang. 2019. Structured knowledge distillation for semantic segmentation. In *Proceedings of the IEEE/CVF Conference on Computer Vision and Pattern Recognition*. 2604–2613.

[18] Aleksander Madry, Aleksandar Makelov, Ludwig Schmidt, Dimitris Tsipras, and Adrian Vladu. 2017. Towards deep learning models resistant to adversarial attacks. *arXiv preprint arXiv:1706.06083* (2017).

[19] Rama Sai Mamidala, Uday Uthkota, Mahamkali Bhavani Shankar, A. Joseph Antony, and A. V. Narasimhadhan. 2019. Dynamic approach for lane detection using google street view and cnn. In *IEEE TENCON*.

[20] Seyed-Mohsen Moosavi-Dezfooli, Alhussein Fawzi, Omar Fawzi, and Pascal Frossard. 2017. Universal adversarial perturbations. In *Proceedings of the IEEE conference on computer vision and pattern recognition*. 1765–1773.

[21] Gaurav Kumar Nayak, Ruchit Rawal, and Anirban Chakraborty. 2022. DAD: Data-Free Adversarial Defense at Test Time. In *Proceedings of the IEEE/CVF Winter Conference on Applications of Computer Vision*. 3562–3571.

[22] Davy Neven, Bert De Brabandere, Stamatios Georgoulis, Marc Proesmans, and Luc Van Gool. 2018. Towards end-to-end lane detection: an instance segmentation approach. In *2018 IEEE intelligent vehicles symposium (IV)*. IEEE, 286–291.

[23] Xingang Pan, Jianping Shi, Ping Luo, Xiaogang Wang, and Xiaoou Tang. 2018. Spatial as deep: Spatial cnn for traffic scene understanding. In *Proceedings of the AAAI Conference on Artificial Intelligence*, Vol. 32.

[24] Jonah Philion. 2019. Fastdraw: Addressing the long tail of lane detection by adapting a sequential prediction network. In *Proceedings of the IEEE/CVF Conference on Computer Vision and Pattern Recognition*. 11582–11591.

[25] Fabio Pizzati, Marco Allodi, Alejandro Barrera, and Fernando García. 2019. Lane detection and classification using cascaded cnns. In *Eurocast*.

[26] Zequn Qin, Huanyu Wang, and Xi Li. 2020. Ultra fast structure-aware deep lane detection. In *European Conference on Computer Vision*. Springer, 276–291.

[27] Eduardo Romera, José M Alvarez, Luis M Bergasa, and Roberto Arroyo. 2017. Erfnet: Efficient residual factorized convnet for real-time semantic segmentation. *IEEE Transactions on Intelligent Transportation Systems* 19, 1 (2017), 263–272.

[28] Jinming Su, Chao Chen, Ke Zhang, Junfeng Luo, Xiaoming Wei, and Xiaolin Wei. 2021. Structure Guided Lane Detection. *arXiv preprint arXiv:2105.05403* (2021).

[29] Christian Szegedy, Wojciech Zaremba, Ilya Sutskever, Joan Bruna, Dumitru Erhan, Ian Goodfellow, and Rob Fergus. 2013. Intriguing properties of neural networks. *arXiv preprint arXiv:1312.6199* (2013).

[30] Lucas Tabelini, Rodrigo Berriel, Thiago M Paixao, Claudine Badue, Alberto F De Souza, and Thiago Oliveira-Santos. 2021. Keep your eyes on the lane: Real-time attention-guided lane detection. In *Proceedings of the IEEE/CVF conference on computer vision and pattern recognition*. 294–302.

[31] Lucas Tabelini, Rodrigo Berriel, Thiago M Paixao, Claudine Badue, Alberto F De Souza, and Thiago Oliveira-Santos. 2021. Polylanenet: Lane estimation via deep polynomial regression. In *2020 25th International Conference on Pattern Recognition (ICPR)*. IEEE, 6150–6156.

[32] Jigang Tang, Songbin Li, and Peng Liu. 2021. A review of lane detection methods based on deep learning. *Pattern Recognition* 111 (2021), 107623.

[33] Tusimple. 2020. Tusimple benchmark. https://github.com/TuSimple/tusimple-benchmark/

[34] Tao Wang, Li Yuan, Xiaopeng Zhang, and Jiashi Feng. 2019. Distilling object detectors with fine-grained feature imitation. In *Proceedings of the IEEE/CVF Conference on Computer Vision and Pattern Recognition*. 4933–4942.

[35] Qizhe Xie, Minh-Thang Luong, Eduard Hovy, and Quoc V Le. 2020. Self-training with noisy student improves imagenet classification. In *Proceedings of the IEEE/CVF conference on computer vision and pattern recognition*. 10687–10698.

[36] Hang Xu, Shaoju Wang, Xinyue Cai, Wei Zhang, Xiaodan Liang, and Zhenguo Li. 2020. Curvelane-nas: Unifying lane-sensitive architecture search and adaptive point blending. In *European Conference on Computer Vision*. Springer, 689–704.

[37] Seungwoo Yoo, Hee Seok Lee, Heesoo Myeong, Sungrack Yun, Hyoungwoo Park, Janghoon Cho, and Duck Hoon Kim. 2020. End-to-end lane marker detection via row-wise classification. In *Proceedings of the IEEE/CVF Conference on Computer Vision and Pattern Recognition Workshops*. 1006–1007.

[38] Feng Zhang, Xiatian Zhu, and Mao Ye. 2019. Fast human pose estimation. In *Proceedings of the IEEE/CVF Conference on Computer Vision and Pattern Recognition*. 3517–3526.

[39] Gongjie Zhang, Kaiwen Cui, Tzu-Yi Hung, and Shijian Lu. 2021. Defect-GAN: High-fidelity defect synthesis for automated defect inspection. In *Proceedings of the IEEE/CVF Winter Conference on Applications of Computer Vision*. 2524–2534.